\documentclass[journal]{IEEEtran}
\ifCLASSINFOpdf
\else
\fi
\usepackage{graphicx}
\usepackage{bbm}
\usepackage{amsmath}
\usepackage{amsthm}
\usepackage{amssymb}
\usepackage{algorithm}
\usepackage{algpseudocode}
\usepackage{cite}
\newtheorem{definition}{Definition}

\DeclareMathOperator{\argmin}{argmin}

\begin{document}
%
\title{Unsupervised Adaptive Semantic Segmentation with Local Lipschitz Constraint}
%
%
%

\author{Guanyu Cai,
			Lianghua He, 
\thanks{G. Cai and L. He are with the Department of Computer Science and Technology, Tongji University, Shanghai 201804, China (email: caiguanyu@tongji.edu.cn; helianghua@tongji.edu.cn).}
}
\maketitle

\begin{abstract}
Recent advances in unsupervised domain adaptation have seen considerable progress in semantic segmentation. Existing methods either align different domains with adversarial training or involve the self-learning that utilizes pseudo labels to conduct supervised training. The former always suffers from the unstable training caused by adversarial training and only focuses on the inter-domain gap that ignores intra-domain knowledge. The latter tends to put overconfident label prediction on wrong categories, which propagates errors to more samples. To solve these problems, we propose a two-stage adaptive semantic segmentation method based on the local Lipschitz constraint that satisfies both domain alignment and domain-specific exploration under a unified principle. In the first stage, we propose the local Lipschitzness regularization as the objective function to align different domains by exploiting intra-domain knowledge, which explores a promising direction for non-adversarial adaptive semantic segmentation. In the second stage, we use the local Lipschitzness regularization to estimate the probability of satisfying Lipschitzness for each pixel, and then dynamically sets the threshold of pseudo labels to conduct self-learning. Such dynamical self-learning effectively avoids the error propagation caused by noisy labels. Optimization in both stages is based on the same principle, i.e., the local Lipschitz constraint, so that the knowledge learned in the first stage can be maintained in the second stage. Further, due to the model-agnostic property, our method can easily adapt to any CNN-based semantic segmentation networks. Experimental results demonstrate the excellent performance of our method on standard benchmarks.
\end{abstract}


%
\IEEEpeerreviewmaketitle

\section{Introduction}
%
%
%
%
\IEEEPARstart{S}{emantic} segmentation is a core problem in computer vision. It aims at assigning each pixel in an image to a semantic category. Recently, deep convolutional neural networks (CNN) \cite{long2015fully,chen2017deeplab,chen2017rethinking,ding2020semantic} 
have brought a remarkable development of semantic segmentation, leading to various applications, such as autonomous
driving \cite{luc2017predicting,zhang2017curriculum}, robotics \cite{milioto2018real,shvets2018automatic}, and medical analysis \cite{li2018h,ronneberger2015u}. However, traditional semantic segmentation requires massive data with pixel-level annotations, which are expensive and labor-intensive. To overcome such limitations, unsupervised domain adaptation (UDA) \cite{pan2009survey} is introduced. It achieves an adaptive semantic segmentation \cite{tsai2018learning,zou2018unsupervised} by transferring knowledge from models that are well trained on annotated data (source domain) to an unlabeled dataset (target domain). Thus, UDA is practical to satisfy performance on the target domain and the efficiency of data collection simultaneously.     

Existing adaptive semantic segmentation methods mainly explore two lines: domain alignment and domain-specific learning. The former one focuses on minimizing the discrepancy between the source and target domain. By aligning different domains in different levels, such as pixel level \cite{wu2018dcan,wu2019ace,hoffman2018cycada,chen2019learning}, feature level \cite{wang2020differential,huang2018domain,yue2019domain,luo2019significance} and semantic level \cite{tsai2018learning,tsai2019domain,wang2019class}, these methods decrease the discrepancy between different domains and perform well on the target domain. However, most of the domain alignment is conducted in adversarial training \cite{tsai2018learning,ganin2016domain,tzeng2017adversarial}. It easily leads to a vanishing gradient problem that leads to unstable training \cite{cai2019unsupervised}. 
Moreover, the domain alignment only learns shared knowledge of different domains to reduce inter-domain gap, whereas the ultimate goal should focus on the target domain. The neglect of intra-domain knowledge limits these methods' performance.
In this work, we explore a non-adversarial domain alignment based on a Lipschitz constraint to avoid the vanishing gradient problem and further exploit intra-domain knowledge with the constraint. Such intra-domain domain alignment is orthogonal to adversarial-training based methods \cite{tsai2018learning,ganin2016domain,tzeng2017adversarial}, thus it can collaborate with them to further enhance the UDA performance.

The domain-specific learning tries to explore domain-specific knowledge in the target domain with a two-stage self-learning. In detail, in the first stage, models that are well trained on the source domain predict pseudo labels of the target domain. Then, by regarding pseudo labels as ground truth data, models are optimized in a supervised manner. Generally, to reach high performance, these methods adopt a domain alignment in the first stage to decrease errors in pseudo labels. In the second stage, they set a threshold manually to avoid the noisy-label problem \cite{natarajan2013learning}, where only pseudo labels with confidence scores higher than the manual threshold are used to conduct a self-training, because noisy labels that are different from the ground truth labels make a model worse after training. Despite achieving excellent performance, these methods confront two problems.
Firstly, the manual threshold is difficult to determine for different settings. For example, for different target domains, because the discrepancy between the source domain and them is different, a suitable threshold for each of them is hard to determine. Meanwhile, a suitable threshold for each category is hard to determine. If a category rarely appears in the source domain and a high threshold is set, the target domain would obtain too few pseudo labels of the category to conduct a self-learning.
Secondly, for most self-learning methods, there is no connection between the domain alignment in the first stage and the self-learning in the second stage. They optimize different objectives in different stages. Typically, adversarial training is utilized to align domains, whereas in stage two, the objective function of adversarial training is moved and a supervised learning is conducted with pseudo labels.
The learned knowledge in stage one is hard to maintain in the second stage. For example, after adversarial learning, we can estimate whether the threshold should be high or low according to the adversarial loss. If the adversarial loss is large, a low threshold is needed to ensure enough pseudo labels to perform a self-learning, because a large adversarial loss means that the model trained on the source domain is hard to generate convincing pseudo labels. Meanwhile, which category is more reliable can be inferred according to the accuracy in the source domain. However, a manual threshold ignores these information. From another aspect, due to the catastrophic forgetting easily appears in continual learning \cite{kirkpatrick2017overcoming}, self-learning in stage two may force a model to forget knowledge learned in stage one if no regularization is introduced.  


In this work, domain alignment and domain-specific learning are both explored to reach better adaptation. To overcome problems in previous adaptive semantic segmentation methods, we propose a Lipschitz-constraint-based method for adaptive semantic segmentation, which introduces a Lipschitz constraint as the unified principle to guide two-stage optimization. Specifically, in stage one, a local Lipschitzness regularization that is conducted in a pixel-wise way is proposed to align different domains without the vanishing gradient problem caused by adversarial learning \cite{cai2019unsupervised}. It is conducted in the target domain to exploit intra-domain knowledge. Further, because it is orthogonal to the adversarial training, combining it with adversarial training together can boost the performance. In stage two, the proposed local Lipschitzness regularization is utilized to estimate the probability of satisfying Lipschitzness for each pixel. The probability is regarded as a dynamic pixel-wise threshold to guide the self-learning. The local Lipschitz constraint is the unified principle that connects the optimization in two stages. It helps transfer knowledge learned in stage one to the second stage and avoids the catastrophic forgetting by introducing the objective in stage one as a regularization term in stage two. Moreover, the proposed method is flexible to build up with any CNN-based semantic segmentation networks and can adopt with other adaptive semantic segmentation methods to boost their performance.

In summary, our contributions are as follows:
\begin{itemize}
    \item Local Lipschitz constraint is the first time, to the best of our knowledge, to be introduced in adaptive semantic segmentation. It align different domains by exploiting intra-domain knowledge. The non-adversarial domain alignment inspires a new promising direction to explore adaptive semantic segmentation.
    \item A dynamic self-learning is reached by using the local Lipschitz constraint to estimate the threshold of pseudo labels. The dynamic self-learning avoids the error propagation caused by noisy labels. The unified principle, i.e, the local Lipschitz constraint, helps connect the two-stage continual learning. It transfers knowledge between different stages and avoids the catastrophic forgetting.
    \item Model agnostic is achieved by our method. It enables a easy adoption to other adaptive semantic segmentation methods without adding extra parameters or requiring specific network architecture.
\end{itemize}


\section{Related Work}
\subsection{Adaptive Semantic Segmentation}
An important challenge in adaptive semantic segmentation is domain alignment. Most adaptive semantic segmentation methods introduce adversarial training \cite{tsai2018learning,ganin2016domain,tzeng2017adversarial} to align the source and target domain in different levels. Several previous works \cite{wu2018dcan,wu2019ace,hoffman2018cycada,chen2019learning} align domains in pixel level. They utilize generative adversarial networks (GANs) to transfer images in the target domain into the visual style of the source domain so that the model trained in the source domain can be applied to the target domain. Some works \cite{wang2020differential,huang2018domain,yue2019domain,luo2019significance} align domains in feature level. In detail, they conduct adversarial training in the middle activation of networks to align the feature distributions. Semantic-level alignment are introduced by \cite{tsai2018learning,ganin2016domain,tzeng2017adversarial}. They focus on aligning semantic output space of different domains. Besides adversarial training, non-adversarial domain alignment \cite{lian2019constructing} is also introduced to achieve a simple alignment without the vanishing gradient problem. 

Instead of domain alignment, some previous works \cite{zou2018unsupervised,zheng2021rectifying,zou2019confidence,shin2020two,pan2020unsupervised} focus on exploring domain-specific knowledge. They always conduct a two-stage self-learning. In the first stage, a model trained on the source domain or trained with a domain alignment method predicts pseudo labels for the target domain. Then, the model is trained in a supervised manner by regarding the pseudo labels as ground truth data in stage two. Because the accuracy of pseudo labels affects the ultimate performance heavily, they typically choose a state-of-the-art domain alignment method in the first stage to achieve strong performance.

\subsection{Lipschitz Constraint}
The Lipschitz constraint is widely explored in GANs, because it helps avoid the gradient vanishing problem and the loss of gradient informativeness \cite{arjovsky2017wasserstein,zhou2019lipschitz,qi2020loss}. They propose different tricks into the discriminator and loss function to ensure the Lipschitz continuity of networks. By modifying the Lipschitz constraint, a local Lipschitz constraint \cite{miyato2018virtual} is introduced to cope with semi-supervised learning. In UDA, a theoretical analysis derives the error bound of UDA based on the Lipschitz constraint \cite{ben2014domain}. Inspired by the theoretical derivation, several methods \cite{shu2018a,mao2019virtual,cai2019learning} apply the Lipschitz constraint to solve the classification problem. Some of them \cite{miyato2018virtual,shu2018a,mao2019virtual} ensure Lipschitz continuity by adding noises in pixel level, whereas the other \cite{cai2019learning} adds noises in feature level. As far as we know, although the Lipschitz constraint demonstrates a promising direction to explore non-adversarial UDA, there is no adaptive semantic segmentation method that introduces the Lipschitz constraint. In this work, we first build up an adaptive semantic segmentation with the Lipschitz constraint.

\subsection{Self-learning}
Self-learning, also called co-training or self-training, is widely studied in semi-supervised learning \cite{chapelle2009semi,zhu2005semi} and unsupervised learning \cite{yarowsky1995unsupervised,lee2017deep}, with various applications, such as computer vision \cite{rosenberg2005semi,xie2020self,roychowdhury2019automatic}, natural language process \cite{maeireizo2004co,riloff2003learning,yarowsky1995unsupervised} and speech understanding \cite{kahn2020self}. The intuition of self-learning is to let a model learn complementary information from other models. Recently, UDA inherits the intuition of self-learning. Some works \cite{french2018selfensembling,saito2018maximum,saito2017asymmetric} use several classifiers to guide each other to cope with UDA classification. Some semantic segmentation UDA works \cite{zou2018unsupervised,zheng2021rectifying,zou2019confidence,shin2020two,pan2020unsupervised} first predict pseudo labels and then use pseudo labels to self-training. A crucial problem for self-learning is the correctness of pseudo labels. When pseudo labels fit the ground truth labels, self-learning enhances the model gradually. Otherwise, such learning makes the model worse and worse.

\section{Methodology}
In this section, we first give a detailed definition of adaptive semantic segmentation and related denotations. Then, in Section \ref{lipschitz}, we give an explanation of why the Lipschitz constraint helps tackle the UDA problems and point out the objectives we need to optimize. Our work explores domain alignment and domain-specific learning simultaneously. In Section \ref{local}, we demonstrate a local Lipschitzness regularization to align different domains, which is the first-stage optimization. In Section \ref{dynamic}, a dynamic self-learning in the second stage is introduced, where each pixel's threshold is dynamically determined to avoid noisy labels based on its local Lipschitzness. In Section \ref{training}, we illustrate the whole training procedure.

\subsection{Problem Definition}
Given a labeled source domain $D_S=\{x_s^i,y_s^i\}_{i=1}^{M}$ and an unlabeled target domain $D_T=\{x_t^j,y_t^j\}_{j=1}^{N}$, where $x$ denotes the input image and $y$ denotes the output semantic segmentation. All $y_t$ in the target domain is inaccessible. We denote $P_S$ over $X_S$ and $P_T$ over $X_T$ as the marginal distributions of source and target domain, respectively. An adaptive semantic segmentation method focuses on learning a projection function $h$ that maps the input image $x$ to $y$. The error of $h$ with respect to $P_T$ is defined as $E_{P_T}(h)=P_T(y_t\neq h(x_t))$. For a practical semantic segmentation adaptation, there exists an optimal function $h^*$ and the optimal error is $E^*(P_T)=min_{h}E_{P_T}(h)$.

\subsection{Lipschitz Constraint for Domain Adaptation}
\label{lipschitz}
A theoretical derivation \cite{ben2014domain} proves that the error of UDA is bounded by a probabilistic Lipschitzness with respect to a binary classification setting. The definition of probabilistic Lipschitzness is defined as follows:
\begin{definition}[Probabilistic Lipschitzness]
Let $\mu:x_t^2\rightarrow\mathbbm{R}^+$ is a divergence metric, and $\Phi:\mathbbm{R}\rightarrow [0,1]$. $f:x_t\rightarrow \mathbbm{R}$ is $\Phi-Lipschitz$ w.r.t. distribution $P_T$ over $x_t$ if, for all $\lambda > 0$:
\begin{align}
P_T[\exists z: |f(x_t)-f(z)|>\lambda\mu(x_t,z)]\leq\Phi(\lambda)
\label{eqlip}
\end{align}
where $z$ is sampled from $P_T$.
\end{definition}

Given a labeled sample batch $S$, for any $x\in S$, $l(x)=y$ denotes the label of $x$ and $N_S(x)$ denotes the nearest neighbor to $x$ in $S$, i.e., $N_S(x)=\argmin_{z\in S_\chi}\mu(x,z)$. Define a hypothesis $\mathbbm{H}$ determined by the Nearest Neighbor algorithm as $\mathbbm{H}(x)=l(N_S(x))$ for all $x\in \chi$. A error bound of UDA is derived as follows:
\begin{align}
\mathop{\mathbb{E}}_{S\sim P^m}[E_{P_T}(\mathbb{H})]\leq2E^*(P_T)+\Phi(\gamma)+\frac{4\gamma\sqrt{d}}{Rm^{\frac{1}{d+1}}}
\label{eqbound}
\end{align} 
where $m$ denotes the size of $S$ that contains points sampled i.i.d. from the input samples according to some distribution $P^m$ and $d$ denotes the dimension of input samples. $R$ is a positive constant defined in weight ratio \cite{ben2014domain}. It is determined by finite samples in the source and target domain.

According to (\ref{eqbound}), the error bound of UDA includes three crucial factors. The first one is the optimal error $E^*(\mathbbm{P_T})$. It is impractical to optimize because it is an ideal value. The second term is the probabilistic Lipschitzness $\Phi(\gamma)$. To reach a tight error bound, it is reasonable to satisfy the Lipschitz constraint to minimize $\Phi(\gamma)$. Thus, in this work, we focus on training an adaptive semantic segmentation model that satisfies the Lipschitz constraint. The third term comprises several constants. It is practical to minimize $m$ and $d$ to obtain a tight error bound, i.e., enlarging the number of samples and decreasing the dimension of samples.

\begin{figure}
    \centering
    \includegraphics[width=1\columnwidth]{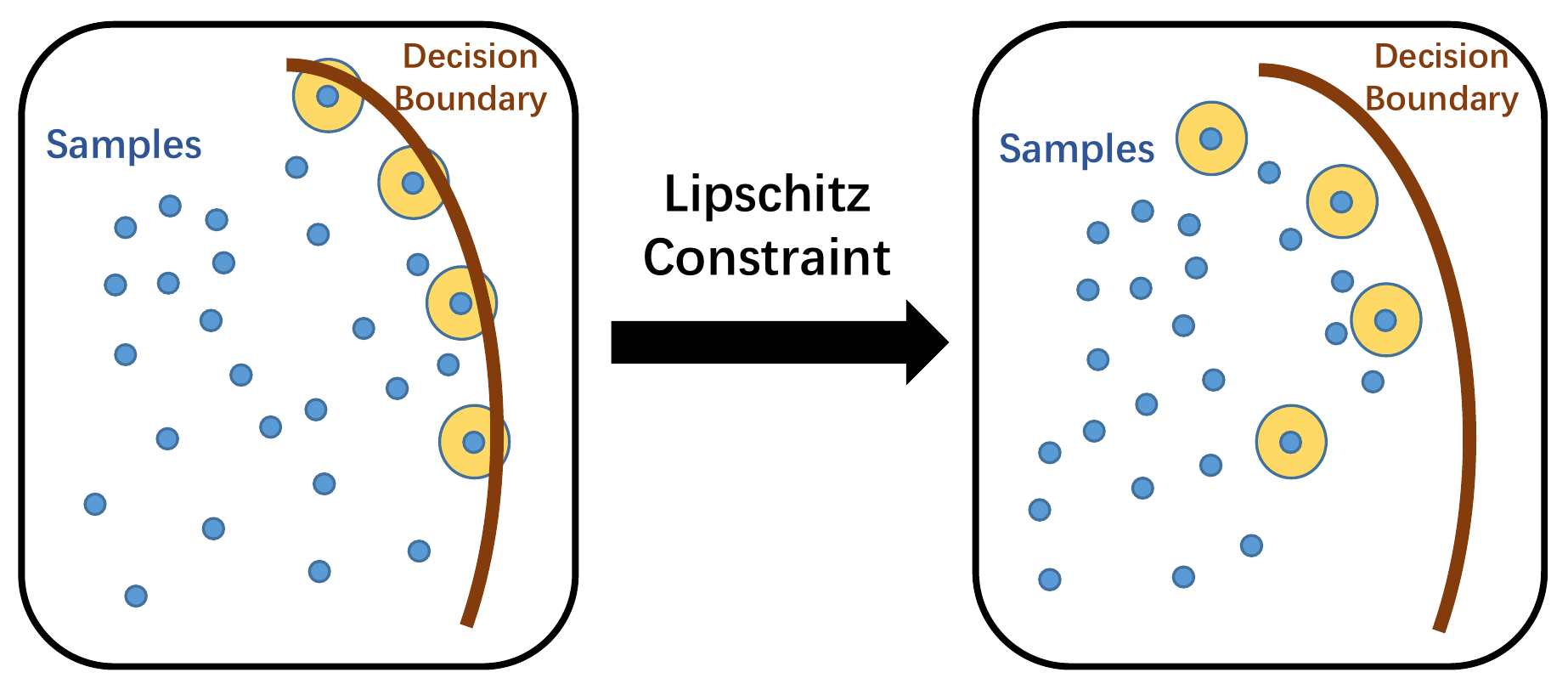}
    \caption{Illustration of the Lipschitz constraint. The Lipschitz constraint forces samples in a local region to output similar results. Thus, samples near the decision boundary would be pushed away from the decision boundary to ensure the output of them to be the same.}
    \label{fig:intuitive}
\end{figure}

An intuitive description of the Lipschitz constraint is shown in Fig. \ref{fig:intuitive}. For samples near the decision boundary, if a domain shift happens, they could easily cross the boundary, thus, are classified into a wrong category. These sensitive samples cause the performance drop when a model that is well trained on the source domain applies to the target domain. The Lipschitz constraint alleviates such performance drop as it pushes sensitive samples away from the decision boundary. Specifically, if a sample is near the decision boundary, samples around it could potentially cross the boundary and their outputs would be different. Because the Lipschitz constraint tries to make these samples' outputs to be the same, these sensitive samples would be pushed away from the boundary to form a robust decision boundary. The robust decision boundary avoids misclassification even a domain shift is introduced.

Although the error bound is derived in a binary classification setting, it indeed inspires us to propose an adaptive semantic segmentation method, because semantic segmentation is also a classification problem that performs in pixel level. In this work, considering the feature that segmentation is a pixel-level problem, we propose a local Lipschitz regularization to satisfy the pixel-level Lipschitz constraint in a neural network. A problem caused by the pixel-level Lipschitz constraint is that a large dimension of image samples leads to a loose error bound as the third term in (\ref{eqbound}) is large. To resolve the conflict, we design an optimization trick to optimize the network in feature level such that the dimension of input samples is greatly reduced, and satisfy the pixel-level Lipschitz constraint at the same time. Thus, both tight error bound and pixel-level Lipschitz constraint are simultaneously satisfied. 

\subsection{Local Lipschitzness Regularization}
\label{local}

\begin{figure}
    \centering
    \includegraphics[width=1\columnwidth]{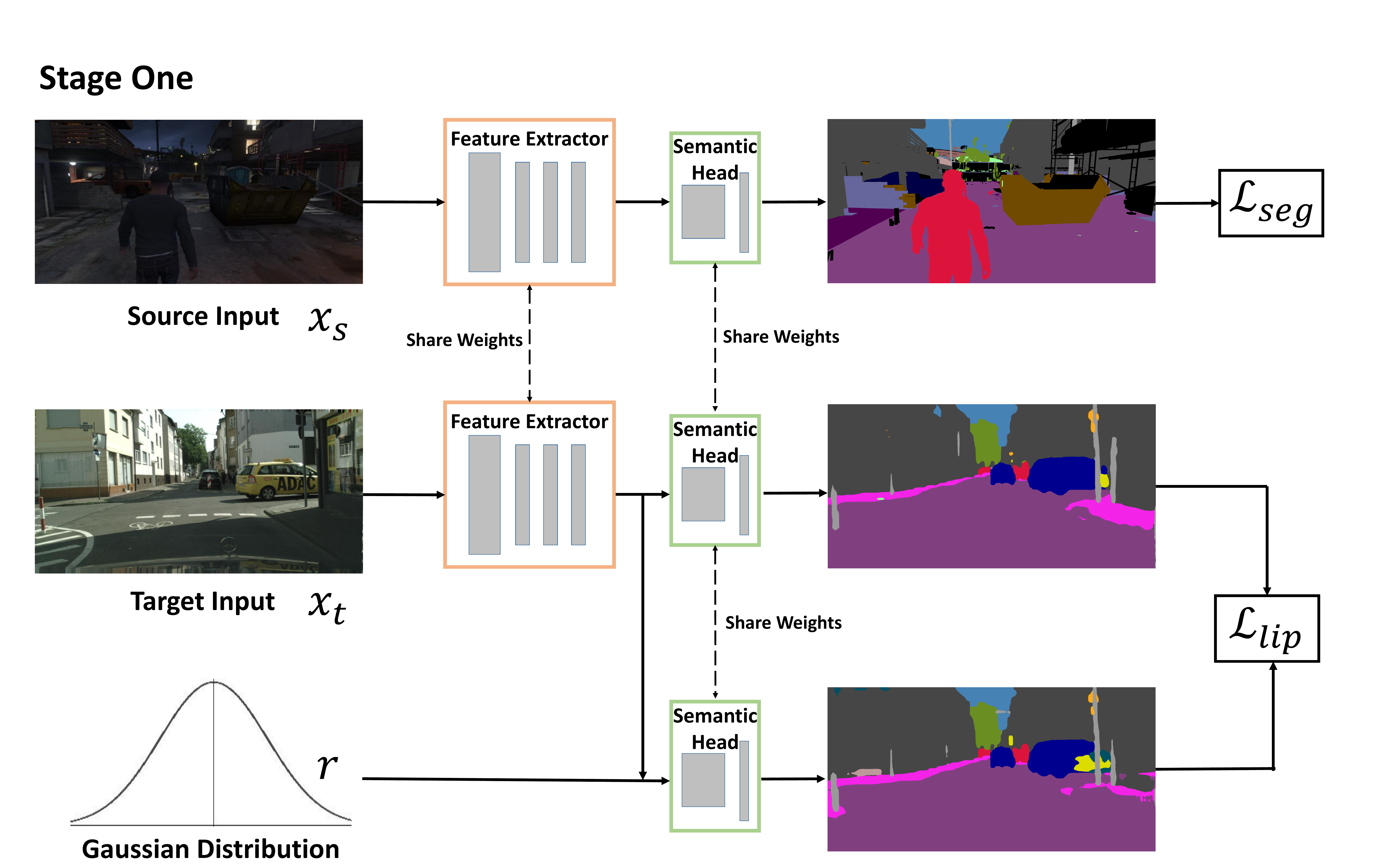}
    \caption{Overview of Stage one. In stage one, samples from the source domain are utilized to conduct a supervised semantic. Meanwhile, the local Lipschitzness regularization forces samples in the target domain to satisfy the Lipschitz constraint.}
    \label{fig:stage1}
\end{figure}

In detail, our method includes two stages, where the first stage takes the charge of aligning different domains and the second stage performs self-learning to further explore domain-specific knowledge. A local Lipschitzness regularization is proposed to satisfy the pixel-level probabilistic Lipschitz constraint. Briefly, a Gaussian noise is added to the middle feature of a network, and we measure the difference between the original semantic output and the one added noises to represent the Lipschitzness of each pixel. To ensure a small dimension of input samples, we add noises to the feature of last convolution layer so that the dimension of input to $f$ in (\ref{eqlip}) is small. This trick regards the downstream layers after the last convolution layer as $f$ in (\ref{eqlip}). The goal of optimization is to make the downstream layers satisfy the Lipschitz constraint. To ensure the pixel-level Lipschitz constraint, the difference of output is calculated in the pixel-level semantic output. In stage one, we regard the local Lipschitzness regularization as a loss function to directly enhance the Lipschitzness. In stage two, the local Lipschitzness regularization is utilized to estimate the probability of satisfying Lipschitzness for each pixel. The probability is regarded as a dynamic pixel-wise threshold to guide the self-learning.

In (\ref{eqlip}), $\Phi(\lambda)$ is the supremum of $P_T[\exists y:|f(x_t)-f(z)|>\lambda\mu(x_t,z)]$. To estimate $P_T[\exists y:|f(x_t)-f(z)|>\lambda\mu(x_t,z)]$, for each $x_t$, we need to sample a bunch of $z$ and calculate $|f(x_t)-f(z)|$. In this work, we propose a local Lipschitzness regularization to avoid such burdensome computation:
\begin{align}
    \label{eqlocal}
    \mathcal{L}_{lip}(x_t,\theta_G, \theta_H)=&D(H(G(x_t)+r),H(G(x_t))),\\\nonumber
    & where\;||r||\leq\epsilon
\end{align}
where $r$ is a noise vector sampled from a Gaussian distribution and $\epsilon$ denotes the maximum norm of $r$. $D$ denotes a divergence metric, such as $L_2$ distance and kullback-leibler divergence. $G$ is the feature extractor parameterized by $\theta_G$ of a semantic segmentation network, e.g., the four basic blocks of ResNet \cite{he2016deep}. $H$ is the semantic head parameterized by $\theta_H$ of a semantic segmentation network, e.g., the interpolation layer, the atrous spatial pyramid pooling module and the prediction layer. $H(G(x))$ performs just like $\mathbbm{H}(x)$ in (\ref{eqbound}).

(\ref{eqlocal}) assumes that every $x$ has a neighborhood $U$ whose norm is constrained by $\epsilon$, and $H$ is optimized to be Lipschitz continuous with respect to $x$ and samples in $U$. Namely, by regarding samples in $U$ as $y$ in  (\ref{eqlip}), $\mathcal{L}_{lip}(x,\theta_G,\theta_C)$ is an approximation of the probabilistic Lipschitzness. $\mathcal{L}_{lip}(x,\theta_G,\theta_C)$ can be calculated in a point-wise way and the computation complexity is greatly reduced, e.g., for each $x$, we can generate many $x+r$ easily to estimate the Lipschitzness instead of sample $y$ from the target domain. Another point needs to pay attention is that both (\ref{eqlip}) and (\ref{eqlocal}) are defined in the target domain. According to (\ref{eqlip}) and (\ref{eqbound}), even though we further ensure the Lipschitzness in the source domain like \cite{shu2018a,mao2019virtual}, it is meaningless for a tight error bound.

In stage one, to align different domains, we combine the local Lipschitzness regularization with a segmentation loss. The overview of the framework in stage one is shown in Fig. \ref{fig:stage1}. The segmentation loss is a common cross-entropy loss for semantic segmentation \cite{long2015fully}. It applies for images in the source domain:
\begin{align}
    \mathcal{L}_{seg}(x_s)=-\sum_{h,w}\sum_{c\in C}y_s^{h,w,c}log(H(G(x_s))^{h,w,c})
    \label{eqseg}
\end{align}
where $h$ denotes the height of output segmentation, $w$ denotes the weight of output segmentation and $C$ denotes the number of categories. Thus, the ultimate objective function in stage one is as follows:
\begin{align}
    \mathcal{L}_1=\mathcal{L}_{seg} + \lambda_1\mathcal{L}_{lip}
    \label{eqstage1}
\end{align}
where $\lambda_1$ is a coefficient to balance the segmentation loss and local Lipschitzness regularization.

The local Lipschitzness regularization focuses on the intra-domain knowledge in the target domain. Intuitively, it makes the classification of each pixel robust enough to avoid an ambiguous output. It is optimized in a non-adversarial training to avoid the vanishing gradient problem. Meanwhile, because the intuition and detailed optimization of it is totally different from adversarial-training based methods \cite{tsai2018learning,tsai2019domain}, it can also collaborate with them to further enhance the UDA performance.

\subsection{Dynamic Self-learning}
\label{dynamic}

\begin{figure}
    \centering
    \includegraphics[width=1\columnwidth]{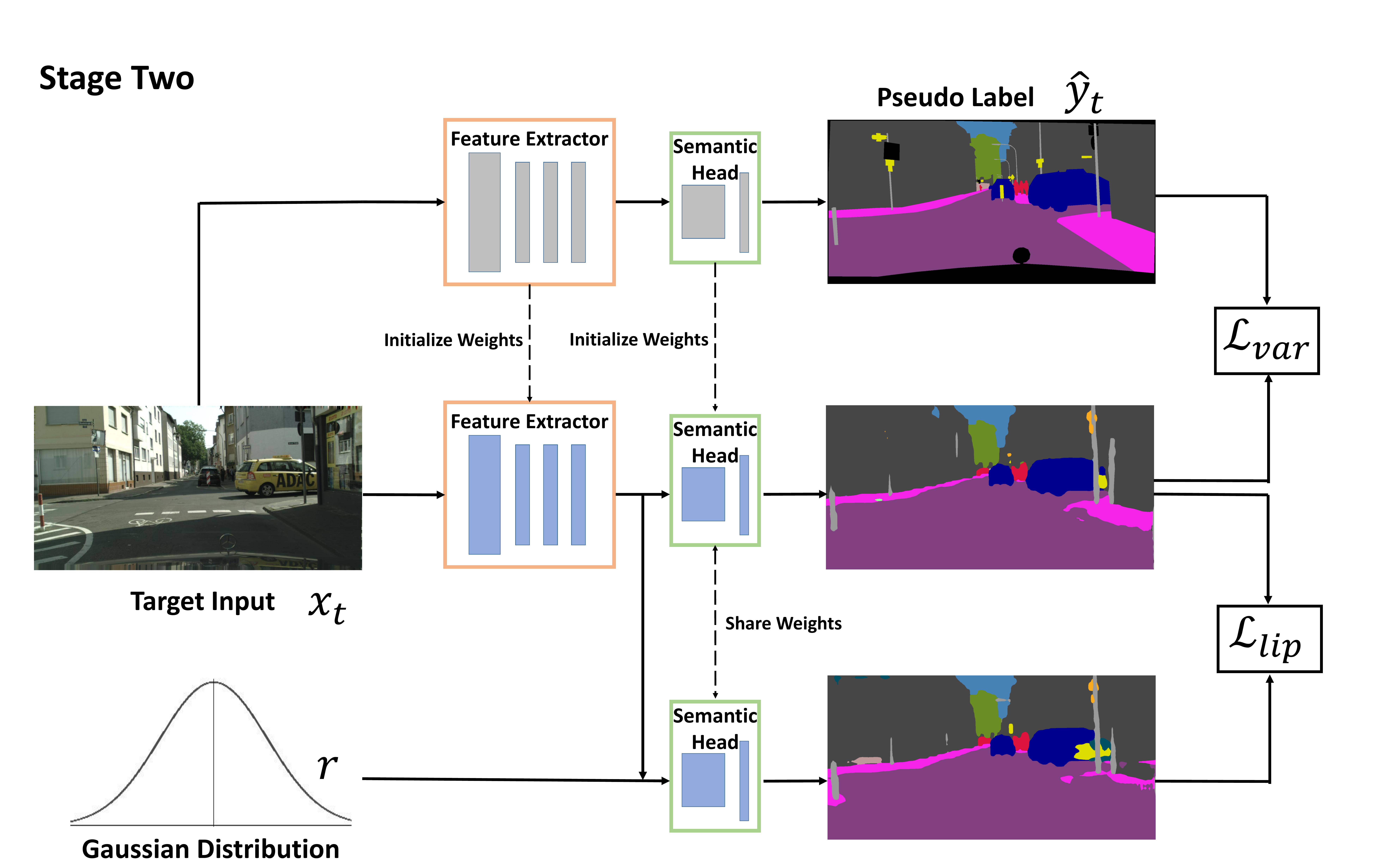}
    \caption{Overview of stage two. In stage two, the local Lipschitzness regularization is used to measure the probability of satisfying Lipschitzness for each pixel. The probability is used to dynamically determine the confidence of pseudo labels. Besides the dynamic self-learning, the local Lipschitzness regularization is also added to the objective function to maintain the knowledge learned in stage one.}
    \label{fig:stage2}
\end{figure}
To explore domain-specific knowledge in the target domain, we introduce a dynamic self-learning, which first utilizes a model to predict pseudo labels of the target domain, and then conduct a supervised training with pseudo labels. In stage two, we can use the model trained in stage one to predict pseudo labels. The conventional self-learning loss of each pixel is as follows:
\begin{align}
    \mathcal{L}_{self}(x_t)=-\hat{y}_t^{h,w,c} log(H(G(x_t))^{h,w,c})
\end{align}
where $\hat{y}_t^{h,w,c}$ is the pseudo label of $x_t$. $h$ denotes the height of output segmentation, $w$ denotes the weight of output segmentation and $C$ denotes the number of categories.

The most challenge issue in self-learning is the noisy labels in pseudo labels that are different from the ground truth labels. If noisy labels are regarded as the ground truth to perform a supervised learning, the model would become worse and worse. To tackle this problem, we utilize the local Lipschitzness regularization to measure the probability of satisfying Lipschitzness for each pixel and use the probability to dynamically set a threshold for each pixel to determine the pseudo label is whether positive or not. 

In detail, we first recall (\ref{eqlocal}) that if a pixel in $x_t$ has a small probability to satisfy Lipschitzness, $\mathcal{L}_{lip}$ would be large for it. Considering that Lipschitzness constrains the error bound in (\ref{eqlip}), we regard pseudo labels of pixels with large $\mathcal{L}_{lip}$ as noisy labels. To avoid the effects of these noisy labels, we can modify the supervised training with pseudo labels by introducing the dynamic thresholds based on $\mathcal{L}_{lip}$:
\begin{align}
    \mathcal{L}_{var}=\sum_{h,w}\sum_{c\in C}\frac{1}{\mathcal{L}_{lip}(x_t)}\mathcal{L}_{self}(x_t)
    \label{eqvar}
\end{align}
where $h$ denotes the height of output segmentation, $w$ denotes the weight of output segmentation and $C$ denotes the number of categories. To stabilize the training, we adopt the policy in \cite{kendall2017uncertainties} that replace $\frac{1}{\mathcal{L}_{lip}(x_t)}$ to $exp(-\mathcal{L}_{lip}(x_t))$. Thus, (\ref{eqvar}) is modified into:
\begin{align}
    \mathcal{L}_{var}=\sum_{h,w}\sum_{c\in C}exp(-\mathcal{L}_{lip}(x_t))\mathcal{L}_{self}(x_t)
    \label{eqmovar}
\end{align}

The proposed self-learning dynamically adjusts the threshold for each pixel. When the pseudo label of a pixel has a large probability to be wrong, the threshold would be large, i.e., its loss would be refined to a small value to decrease its impact on $\theta_G$ and $\theta_C$. Such dynamical threshold is more reasonable than previous manual threshold \cite{zou2018unsupervised}, because in different settings, such as different datasets, different categories and different locations, the threshold should be adaptively chosen. It is burdensome for people to determine the threshold for different settings accurately, whereas our method uses the Lipschitz constraint to dynamically determine it. It is more convenient and has a convincing theoretical support. Another advantage of the proposed self-training is that no extra parameters or modules are added. Different from \cite{yu2019robust} that introduces the extra branches, and \cite{zheng2021rectifying} that requires two semantic heads, our method can be easily adapt to any CNN-based semantic segmentation networks without adding extra parameters. This makes it more practical. We verify the model-agnostic property of the proposed dynamic self-learning with an ablation study in Section \ref{ablation}.

Another problem in previous self-learning methods \cite{zou2018unsupervised,zou2019confidence} is that they ignore the objective function in state one. It is easy to leads the catastrophic forgetting \cite{kirkpatrick2017overcoming} that forgets knowledge learned in stage one if no regularization is introduced. To consolidate knowledge learned in stage one, we add $\mathcal{L}_{lip}$ as regularization to (\ref{eqmovar}) to avoid the catastrophic forgetting:
\begin{align}
    \mathcal{L}_2=\mathcal{L}_{val} + \lambda_2\mathcal{L}_{lip}
    \label{eqstage2}
\end{align}
where $\lambda_2$ is a coefficient to balance the self-learning loss and the local Lipschitzness regularization. The overview of the framework in stage two is shown in Fig. \ref{fig:stage2}.

\subsection{Training Procedure}
\label{training}
By combining the first and second stages, the whole training procedure is formed. First, we train the model to optimize the objective function in (\ref{eqstage1}) to align different domains with the local Lipschitzness regularization. Because the optimization is non-adversarial, the training procedure gets rid of the vanishing gradient problem \cite{cai2019unsupervised}. Meanwhile, the local Lipschitz regularization focuses on the intra-domain knowledge and adversarial training tries to learn the inter-domain knowledge, thus, our method can collaborate with adversarial training to further boost the UDA performance.
Second, after predicting pseudo labels of the target domain with the model trained in stage one. We further train the model to optimize the objective function in (\ref{eqstage2}) to conduct the dynamic self-learning with pseudo labels. 

Another advantage of the proposed method is that it can easily adapt to any CNN-based semantic segmentation backbones and any other adaptive semantic segmentation methods. When calculating $\mathcal{L}_{lip}$, noises are added to the middle feature of a network. This operation has no requirement about the network architecture. Thus, our method can adapt to any CNN-based semantic segmentation backbones. Further, because the optimization in stage two only needs pseudo labels and our method has no requirement about the network architecture, it can adapt to any other adaptive semantic segmentation methods only if they could predict pseudo labels. The whole training procedure is shown in Algorithm \ref{alg:1}.

\begin{algorithm}[H]
\caption{Training procedure of the proposed method.}
\label{alg:1}
\begin{algorithmic}
\State Initialize $\theta_G$ and $\theta_C$, the iteration number of stage one $T_1$ and the iteration number of stage two $T_2$.
\State\textbf{Input:} The source domain dataset $\{x_s^i,y_s^i\}_{i=1}^{M}$ and the target domain dataset $\{x_t^j\}_{j=1}^{N}$.
\State\textbf{Stage one:}
\For{iteration=1,$T_1$}
\State Input $x_s^i,y_s^i$ to $G$ and $C$ to compute the segmentation loss $\mathcal{L}_{seg}$ according to (\ref{eqseg});
\State Input $x_t^j$ to $G$ and $C$ to compute the local Lipschitzness regularization $\mathcal{L}_{lip}$ according to (\ref{eqlocal});
\State Optimize $\mathcal{L}_1=\mathcal{L}_{seg}+\lambda_1\mathcal{L}_{lip}$.
\EndFor

\State\textbf{Stage two:}
\State Predict pseudo labels of the target domain $\{\hat{y}_t^j\}_{j=1}^{N}$.
\For{iteration=1,$T_2$}
\State Input $x_t^j$ to $G$ and $C$ to compute the local Lipschitzness regularization $\mathcal{L}_{lip}$ according to (\ref{eqlocal});
\State Input $x_t^j,\hat{y}_t^j$ to $G$ and $C$ to compute the self-learning loss according to (\ref{eqmovar});
\State Optimize $\mathcal{L}_2=\mathcal{L}_{val} + \lambda_2\mathcal{L}_{lip}$.
\EndFor
\State\textbf{Return:} $\theta_G$ and $\theta_C$.
\end{algorithmic}
\end{algorithm}

\section{Experiment}
To verify the effectiveness of our method, we perform several experiments. First, we introduce the datasets. Then, the implementation details are illustrated. Third, we show the results on different datasets. Visualization and the ablation study are also demonstrated in this section. Specifically, we denote the first stage as {\bf L}ipschitz-{\bf C}onstraint-based {\bf D}omain {\bf A}daptation (LCDA) and the second stage as {\bf L}ipschitz-{\bf C}onstraint-based {\bf R}e{\bf F}inement (LCRF).

\subsection{Datasets and Metric}
\textbf{Datasets.} We conduct experiments on three standard benchmarks: GTA5 \cite{richter2016playing}, SYNTHIA \cite{ros2016synthia} and Cityscapes \cite{cordts2016cityscapes}. GTA5 is collected from a video game. It contains 24,966 images. SYNTHIA is rendered from a virtual city and is annotated with pixel-level segmentation. It contains 9,400 images. Cityscapes is a realistic dataset, which collects street-view scenes from 50 different cities. It contains 2,975 images for training and 500 images for validation. Considering we want to adapt models trained in a synthetic environment to a realistic one in practice, we form two domain adaptation tasks: GTA5 $\rightarrow$ Cityscapes and SYNTHIA $\rightarrow$ Cityscapes. In detail, the notation A $\rightarrow$ B means that A is the source domain and B is the target domain. Our goal is to adapt the model well-trained on the source domain to the target one without annotations of the target domain.

\textbf{Metric.} Following previous adaptive semantic segmentation methods \cite{zou2018unsupervised,zheng2021rectifying,zou2019confidence}, we report per-category IoU and mean IoU (mIoU) over all categories. For SYNTHIA $\rightarrow$ Cityscapes, because the categories in the source domain are less than these in the target domain, we report the results of 13 categories and 16 categories that are common parts of the source and target domain.

\subsection{Implementation Details}
\label{implementation}
\textbf{Backbone.} Although our method can adapt to any CNN-based semantic segmentation network architectures, to ensure a fair comparison, we deploy the widely-used Deeplab-v2 \cite{chen2017deeplab} as the backbone, which is based on ResNet101 \cite{he2016deep}. Meanwhile, following most previous adaptive semantic segmentation methods \cite{zou2018unsupervised,zou2019confidence,tsai2018learning}, we add an auxiliary semantic head to boost the performance. The auxiliary semantic head has a similar design to the primary one where an atrous spatial pyramid pooling module and a dense layer are included. The auxiliary semantic head is added after \textit{res4b22} layer. The primary semantic head is added after \textit{res5c} layer.

\textbf{Training Setting.} The input image is resized to 1280$\times$640 with scale jittering from $[0.8,1.2]$, and then is randomly cropped to 512$\times$256 for training. Batchsize is set to 2 for the first stage and is changed to 9 for the second stage. Learning rate is set to 0.0002 for stage one and is set to 0.0001 for stage two. In the first stage, $\lambda_1$ is set to 0.2 for the auxiliary head and is set to 1 for the primary head. In the second stage, $\lambda_2$ is set to 1 for both semantic heads. Similar to \cite{zheng2021rectifying}, we deploy the learning rate policy by multiplying the factor $(1-\frac{iter}{total-iter})$. The total iteration is to 100$k$ steps and the early-stop strategy is used. The training procedure is stopped after 40$k$ steps. As for the divergence metric in (\ref{eqlocal}), we can choose many metrics, such as $L_2$ distance and $L_1$ distance. Generally, we use kullback-leibler divergence. Thus, (\ref{eqlocal}) is detailed as:
\begin{align}
     \label{eqmolocal}
    \mathcal{L}_{lip}(x_t)=&\sum_{h,w,c}H(G(x_t))^{h,w,c}log\frac{H(G(x_t))^{h,w,c}}{H(G(x_t)+r)^{h,w,c}},\\\nonumber
    & where\;||r||\leq\epsilon
\end{align}
When multiple semantic heads are adopted in the network architecture, we follow the ensemble trick in \cite{zheng2021rectifying} that combines the output of all semantic output as the final inference output. Specifically, if two semantic heads are used, the final output equals the output of the primary head plus the output of the auxiliary head multiplying 0.5.

{\bf Models.} To show the adaptability of the proposed method, besides implementing LCDA and LCRF, we adapt LCDA and LCRF with several adaptive semantic segmentation methods. In detail, to show that the local Lipschitzness regularization can cooperated with adversarial learning, we implement a model named LCDA+Adv that adds the objective function in AdaptSeg \cite{tsai2018learning} to $\mathcal{L}_1$. Further, to show the superiority of LCRF compared with the manual threshold for pseudo-label learning, we implement a model named LCDA+Adv+Pseudo that utilizes a manual threshold \cite{zou2018unsupervised} to conduct pseudo-label learning over LCDA+Adv. Instead, LCDA+Adv+LCRF uses the proposed LCRF to refine the results of LCDA+Adv. To verify that LCRF generalizes well with other adaptive semantic segmentation models, we adapt it with MRNET \cite{ijcai2020-150}, denoted as MRNET+LCRF. 

\subsection{Results}

{\bf GTA5 $\rightarrow$ Cityscapes.} Results of GTA5 $\rightarrow$ Cityscapes experiment are shown in Table \ref{tab:gta}. We compare the proposed method with other recent adaptive semantic segmentation methods. Results of all models are based on the same backbone, i.e., DeepLabv2 \cite{chen2017deeplab}. These methods can be divided into two categories, i.e., domain-alignment-based and pseudo-label-based methods. In detail, the former includes AdaptSeg \cite{tsai2018learning}, SIBAN \cite{luo2019significance}, CLAN \cite{luo2019taking}, APODA \cite{yang2020adversarial}, PatchAlign \cite{tsai2019domain}, AdvEnt \cite{tsai2019domain}, and MRNET \cite{ijcai2020-150}. The latter includes CBST \cite{zou2018unsupervised}, LRENT \cite{zou2019confidence}, MRL2 \cite{zou2019confidence}, MRENT \cite{zou2019confidence}, MRKLD \cite{zou2019confidence}, MRNET+Pseudo \cite{ijcai2020-150} and RPLUE \cite{zheng2021rectifying}. 

Overall,  MRNET+LCRF and RPLUE are the top two methods on the task. MRNET+LCRF achieves the result 49.9\% mIoU that ranks second among all methods. LCDA obtains 44.0\% mIoU that improves the method without adaptation by 6.7\% mIoU. The obvious improvement verifies the effectiveness of the proposed local Lipschitzness regularization. After adding adversarial training into LCDA, the performance of LCDA+Adv increases to 45.2\%. The improvement shows that the Local Lipschitzness regularization, which is a non-adversarial domain alignment, further enhances adversarial training. Self-learning-based methods also refine the results of domain-alignment-based methods. Specifically, LCDA+Adv+Pseudo, which adopts pseudo-label learning with a manual threshold, improves the result of LCDA+Adv by 2.9\% mIoU. LCRF performs better than Pseudo. It achieves 3.9\% mIoU improvement over LCDA+Adv. The results on GTA5 $\rightarrow$ Cityscapes show that LCDA and LCRF can effectively transfer the knowledge of the source domain to the target one. 

{\bf SYNTHIA $\rightarrow$ Cityscapes.} Results are shown in Table \ref{tab:syn}. We compare the proposed method with other recent adaptive semantic segmentation methods. For a fair comparison, we report all results based on the same backbone, i.e., DeepLabv2 \cite{chen2017deeplab}. These methods can be divided into two categories, i.e., domain-alignment-based and pseudo-label-based methods. In detail, the former includes AdaptSeg \cite{tsai2018learning}, SIBAN \cite{luo2019significance}, CLAN \cite{luo2019taking}, APODA \cite{yang2020adversarial}, PatchAlign \cite{tsai2019domain}, AdvEnt \cite{tsai2019domain} CCM \cite{li2020content} and MRNET \cite{ijcai2020-150}. The latter includes CBST \cite{zou2018unsupervised}, LRENT \cite{zou2019confidence}, MRL2 \cite{zou2019confidence}, MRENT \cite{zou2019confidence}, MRKLD \cite{zou2019confidence}, MRNET+Pseudo \cite{ijcai2020-150} and RPLUE \cite{zheng2021rectifying}. The mean IoU of 16 categories is denoted as mIoU while the mean IoU of 13 categories is denoted as mIoU$^{*}$.

Overall, MRNET+LCRF performs the best on the task. MRNET+LCRF arrives at the state-of-the-art result 48.2\% mIoU$^{*}$ among 16 categories that surpasses other methods. It also achieves 54.9\% mIoU among 13 categories that ranks the first. LCDA achieves 40.9\% mIoU and 47.4\% mIoU$^{*}$ that gains 5.7\% mIoU and 7.0\% mIoU$^{*}$ improvement over the model without adaptation. It verifies the effectiveness of the proposed local Lipschitzness regularization. By adding adversarial training to LCDA, LCDA+Adv gains 1.6\% mIoU and 1.8\% mIoU$^{*}$ improvement over LCDA. Such improvement indicates that the local Lipschitzness regularization can cooperate with adversarial training and further boost the performance of LCDA. Comparing with LCDA+Adv, LCDA+Adv+Pseudo that adopts pseudo-label learning with a manual threshold brings 3.8\% mIoU and 3.8\% mIoU$^{*}$ improvement. The dynamic self-learning, i.e., LCRF, gains 4.4\% mIoU and 4.8\% mIoU$^{*}$ improvement over LCDA+Adv, which performs better than the pseudo-label learning with a manual threshold. The comparison between LCDA+Adv+Pseudo and LCDA+Adv+LCRF shows the superiority of our proposed dynamic self-training. Besides, MRNET+LCRF yields the competitive performance in terms of the per-category IoU. It ranks the first in 6 categories. In several categories that are hard to adapt, such as {\tt wall} and {\tt traffic Light}, it performs much better than other methods.

\renewcommand\arraystretch{1.0}
\begin{table*}
\begin{center}
\resizebox{\linewidth}{32mm}{\begin{tabular}{|l|ccccccccccccccccccc|c|}
\hline
Method & Road & SW & Build & Wall & Fence & Pole & TL & TS &Veg &Terrian & Sky & PR & Rider & Car & Truck & Bus & Train & Motor & Bike & mIoU\\
\hline\hline
Source & 75.8 & 16.8& 77.2 & 12.5 & 21.0 & 25.5 & 30.1 & 20.1 & 81.3 & 24.6 & 70.3 & 53.8 & 26.4 & 49.9 & 17.2 & 25.9 & 6.5 & 25.3 & 36.0 & 36.6\\
AdaptSeg \cite{tsai2018learning} & 86.5 & 36.0 & 79.9 & 23.4 & 23.3 & 23.9 & 35.2 & 14.8 & 83.4 & 33.3 & 75.6 & 58.5 & 27.6 & 73.7 & 32.5 & 35.4 & 3.9 & 30.1 & 28.1 & 42.4\\
SIBAN \cite{luo2019significance} & 88.5 & 35.4& 79.5 & 26.3 &  24.3 & 28.5 &  32.5 & 18.3 &  81.2 & 40.0 & 76.5 & 58.1 &  25.8 & 82.6 & 30.3 & 34.4 & 3.4 & 21.6 & 21.5 & 42.6\\
CLAN \cite{luo2019taking} & 87.0 &27.1& 79.6 &  27.3 &   23.3 & 28.3 &  35.5 & 24.2 &  83.6 & 27.4 & 74.2 & 58.6 &  28.0 &  76.2 &  33.1 &  36.7 &  6.7 &  31.9 & 31.4 & 43.2\\
APODA \cite{yang2020adversarial} & 85.6 &32.8& 79.0 &  29.5 &   25.5 & 26.8 &  34.6 &  19.9 &  83.7 & {\bf 40.6} & 77.9 & 59.2 &  28.3 &  84.6 &  34.6 &  49.2& 8.0 &  32.6 & 39.6 & 45.9\\
PatchAlign \cite{tsai2019domain} & 92.3 &51.9& 82.1 & 29.2 &  25.1 & 24.5 &  33.8& 33.0 & 82.4 &  32.8 & 82.2 & 58.6 &  27.2 &  84.3 & 33.4 &  46.3 &  2.2 &   29.5 & 32.3 & 46.5\\
AdvEnt \cite{tsai2019domain} & 89.4 &33.1& 81.0 & 26.6 &  {\bf 26.8} & 27.2 &  33.5& 24.7 &  83.9 & 36.7 & 78.8 & 58.7 &  30.5 &  84.8 &38.5 &  44.5 &  1.7 &   {\bf 31.6} & 32.4 & 45.5\\
\hline
Source & 71.3 &  19.2& 69.1 &  18.4 & 10.0 & 35.7 & 27.3 & 6.8 & 79.6 & 24.8 & 72.1 & 57.6 &  19.5 & 55.5 & 15.5 & 15.1 & 11.7 & 21.1 & 12.0 & 33.8\\
CBST \cite{zou2018unsupervised} & 91.8 &  53.5& 80.5 & 32.7 & 21.0 & 34.0 & 28.9 &20.4 & 83.9 & 34.2 & 80.9 & 53.1 &  24.0 & 82.7 & 30.3 & 35.9 & 16.0 & 25.9 &42.8 & 45.9\\
LRENT \cite{zou2019confidence}& 91.8& 53.5& 80.5& 32.7& 21.0& 34.0& 29.0& 20.3& 83.9& 34.2& 80.9& 53.1& 23.9& 82.7& 30.2& 35.6& 16.3& 25.9& 42.8 &45.9 \\
MRL2 \cite{zou2019confidence} & {\bf 91.9} & 55.2& 80.9 & 32.1 & 21.5 & 36.7 & 30.0 &19.0 &84.8 & 34.9 & 80.1 & 56.1 &  23.8 & 83.9 & 28.0 & 29.4 & 20.5 & 24.0 &40.3 & 46.0\\
MRENT \cite{zou2019confidence}& 91.8& 53.4& 80.6& 32.6& 20.8& 34.3& 29.7& 21.0& 84.0& 34.1& 80.6& 53.9& 24.6& 82.8& 30.8& 34.9& 16.6& 26.4& 42.6& 46.1 \\
MRKLD \cite{zou2019confidence}& 91.0& {\bf 55.4}& 80.0& 33.7& 21.4& 37.3& 32.9& 24.5& 85.0& 34.1& 80.8& 57.7& 24.6& 84.1& 27.8& 30.1& {\bf 26.9}& 26.0 &42.3 & 47.1\\
\hline
Source& 51.1& 18.3& 75.8& 18.8& 16.8& 34.7& 36.3& 27.2& 80.0& 23.3& 64.9& 59.2& 19.3& 74.6& 26.7& 13.8& 0.1& 32.4& 34.0& 37.2\\
MRNET \cite{ijcai2020-150}& 89.1& 23.9& 82.2& 19.5& 20.1& 33.5& 42.2& 39.1& {\bf 85.3}& 33.7& 76.4& 60.2& 33.7& 86.0& 36.1& 43.3& 5.9& 22.8& 30.8& 45.5\\
MRNET+Pseudo \cite{ijcai2020-150}& 90.5& 35.0& 84.6& 34.3& 24.0& 36.8& 44.1& 42.7& 84.5& 33.6& {\bf 82.5}& 63.1& 34.4& 85.8& 32.9& 38.2& 2.0& 27.1& 41.8& 48.3 \\
RPLUE \cite{zheng2021rectifying}& 90.4& 31.2& {\bf 85.1}& 36.9& 25.6& {\bf 37.5}& {\bf 48.8}& {\bf 48.5}& {\bf 85.3}& 34.8& 81.1& {\bf 64.4}& {\bf 36.8}& 86.3& 34.9& {\bf 52.2}& 1.7& 29.0& 44.6& {\bf 50.3}\\
\hline
Source& 65.9& 21.8& 71.6& 30.0& 25.4& 29.4& 30.3& 14.5& 83.1& 27.8& 74.6& 59.5& 11.3& 75.7& 37.0& 19.8& 2.0& 25.6& 2.86& 37.3\\
LCDA & 89.9& 25.2& 80.6& 22.7& 18.3& 36.2& 39.0& 34.5& 84.3& 28.6& 74.0& 61.5& 28.5& 85.4& 30.8& 43.4& 0.4& 20.6& 31.7& 44.0\\
LCDA+Adv & 90.1& 35.6& 82.3& 27.2& 23.9& 36.1& 38.8& 30.2& 84.1& 34.0& 70.8& 61.6& 31.6& 85.1& 34.1& 40.2& 3.4& 20.7& 29.7& 45.2\\
LCDA+Adv+Pseudo & 91.3& 44.1& 83.6& 32.4& 23.1& 34.6& 42.8& 41.5& 81.9& 36.9& 76.7& 59.8& 33.7& 85.5& 32.9& 46.5& 4.3& 20.9& 41.9 & 48.1\\
LCDA+Adv+LCRF & 91.6& 41.0& 84.0& 36.4& 24.8& 35.0& 43.2& 39.4& 82.8& 34.3& 76.9& 62.3& 36.0& {\bf 86.9}& {\bf 40.8}& 48.1& 0.2& 23.5& 46.2& 49.1\\
MRNET+LCRF & 90.9& 39.8& 84.4& {\bf 40.4}& 24.8& 34.9& 48.0& 47.8& 84.0& 36.2& {\bf 82.5}& 60.7& 33.8& 84.3& 34.8& 41.3& 0.1& 31.5& {\bf 46.4} & 49.9\\
\hline
\end{tabular}}
\end{center}
\caption{Results of GTA5 $\rightarrow$ Cityscapes. Both per-category IoU and mIoU are presented. The results of other works are cited from their papers. The best result in each column is in {\bf BOLD}.}
\label{tab:gta}
\end{table*}

\renewcommand\arraystretch{1.0}
\begin{table*}
\begin{center}
\resizebox{\linewidth}{34mm}{\begin{tabular}{|l|cccccccccccccccc|c|c|}
\hline
Method & Road & SW & Build & Wall$^*$ & Fence$^*$ & Pole$^*$ & TL & TS &Veg & Sky & PR & Rider & Car & Bus & Motor & Bike & mIoU$^*$&mIoU\\
\hline\hline
Source &55.6& 23.8& 74.6& $-$ & $-$ & $-$ & 6.1& 12.1& 74.8& 79.0& 55.3& 19.1& 39.6& 23.3& 13.7& 25.0& 38.6& $-$\\
AdaptSeg \cite{tsai2018learning}& 84.3& 42.7& 77.5& $-$& $-$& $-$& 4.7& 7.0& 77.9& 82.5& 54.3& 21.0& 72.3& 32.2& 18.9& 32.3& 46.7& $-$\\
SIBAN \cite{luo2019significance}& 82.5& 24.0& 79.4& $-$& $-$ &$-$& 16.5& 12.7& 79.2& 82.8& 58.3& 18.0& 79.3& 25.3& 17.6& 25.9& 46.3& $-$ \\
CLAN \cite{luo2019taking}& 81.3& 37.0& 80.1& $-$& $-$& $-$& 16.1& 13.7& 78.2& 81.5& 53.4& 21.2& 73.0& 32.9& 22.6& 30.7& 47.8& $-$\\
APODA \cite{yang2020adversarial}& 86.4& 41.3& 79.3& $-$& $-$& $-$& 22.6& 17.3& 80.3& 81.6& 56.9& 21.0& 84.1& {\bf 49.1}& 24.6& 45.7& 53.1& $-$\\
PatchAlign \cite{tsai2019domain}& 82.4& 38.0& 78.6& 8.7& 0.6& 26.0& 3.9& 11.1& 75.5& 84.6& 53.5& 21.6& 71.4& 32.6& 19.3& 31.7& 46.5& 40.0\\
AdvEnt \cite{tsai2019domain}& 85.6& 42.2& 79.7& 8.7& 0.4& 25.9& 5.4& 8.1& 80.4& 84.1& 57.9& 23.8& 73.3& 36.4& 14.2& 33.0& 48.0& 41.2\\
CCM \cite{li2020content}& 79.6& 36.4& 80.6& 13.3& 0.3& 25.5& 22.4& 14.9& 81.8& 77.4& 56.8& 25.9& 80.7& 45.3& {\bf 29.9}& 52.0& 52.9& 45.2 \\
\hline
Source& 64.3& 21.3& 73.1& 2.4& 1.1& 31.4& 7.0& 27.7& 63.1& 67.6& 42.2& 19.9& 73.1& 15.3& 10.5& 38.9& 34.9&40.3\\
CBST \cite{zou2018unsupervised}& 68.0& 29.9& 76.3& 10.8& 1.4& 33.9& 22.8& 29.5& 77.6& 78.3& 60.6& 28.3& 81.6& 23.5& 18.8& 39.8& 48.9& 42.6\\
LRENT \cite{zou2019confidence}&65.6& 30.3& 74.6& 13.8& 1.5& 35.8& 23.1& 29.1& 77.0& 77.5& 60.1& 28.5& 82.2& 22.6& 20.1& 41.9& 48.7& 42.7\\
MRL2 \cite{zou2019confidence}& 63.4& 27.1& 76.4& 14.2& 1.4 &35.2& 23.6& 29.4& 78.5& 77.8& 61.4& {\bf 29.5}& 82.2& 22.8& 18.9& 42.3& 48.7& 42.8\\
MRENT \cite{zou2019confidence}&69.6& 32.6& 75.8& 12.2& 1.8 &35.3& 23.3& 29.5& 77.7& 78.9& 60.0 &28.5& 81.5& 25.9& 19.6& 41.8& 49.6& 43.4\\
MRKLD \cite{zou2019confidence}&67.7& 32.2& 73.9& 10.7& 1.6& 37.4& 22.2& {\bf 31.2}& 80.8& 80.5& 60.8& 29.1& 82.8& 25.0& 19.4& 45.3& 50.1& 43.8\\
\hline
Source& 44.0& 19.3& 70.9& 8.7& 0.8& 28.2& 16.1& 16.7& 79.8 &81.4 &57.8& 19.2& 46.9 &17.2 &12.0& 43.8& 40.4& 35.2\\
MRNET \cite{ijcai2020-150}&82.0 &36.5& 80.4& 4.2& 0.4 &33.7& 18.0& 13.4& 81.1& 80.8& 61.3& 21.7& 84.4& 32.4& 14.8& 45.7 &50.2& 43.2\\
MRNET+Pseudo \cite{ijcai2020-150}& 83.1& 38.2& 81.7& 9.3& 1.0& 35.1& 30.3& 19.9& {\bf 82.0}& 80.1& 62.8& 21.1& 84.4& 37.8& 24.5& 53.3& 53.8& 46.5\\
RPLUE \cite{zheng2021rectifying}&87.6& {\bf 41.9}& 83.1& 14.7& 1.7& {\bf 36.2}& 31.3& 19.9& 81.6& 80.6& 63.0& 21.8& {\bf 86.2}& 40.7& 23.6& 53.1 &{\bf 54.9}& 47.9\\
\hline
Source&44.0& 19.3& 70.9& 8.7& 0.8& 28.2& 16.1& 16.7& 79.8& 81.4& 57.8& 19.2& 46.9& 17.2& 12.0& 43.8& 40.4& 35.2\\
LCDA &60.6& 24.6& 77.8& 4.1 & 0.3 & 34.4 & 21.3& 14.1& 81.1& 83.2& 59.8& 24.5& 74.4& 32.6& 13.6& 49.0& 47.4& 40.9\\
LCDA+Adv &65.5& 26.0& 80.1& 6.8 & 0.3 & 33.7 & 21.6& 12.9& 81.9& {\bf 84.7}& 61.1& 24.5& 83.1& 36.2& 12.4& 49.8& 49.2& 42.5\\
LCDA+Adv+Pseudo &70.2& 28.6& 81.6& 13.8 & 0.9 & 36.0 & 26.9& 16.1& 80.8& 80.9& {\bf 65.0}& 28.8& 85.3& 45.2& 26.0& 53.9& 53.0&46.3 \\
LCDA+Adv+LCRF &72.5& 29.9& 81.7& 11.2 & 0.9 & 36.1 & 35.1& 18.7& 81.9& 85.5& 63.2& 23.7& 86.0& 47.2& 21.8& {\bf 54.3}& 54.0 & 46.9\\
MRNET+LCRF &{\bf 87.9}& 41.4& {\bf 83.6}5& {\bf 21.6} & {\bf 4.2} & 32.8 & {\bf 36.1}& 24.7& 79.2& 78.2& 59.4& 20.8& {\bf 86.2}& 37.8& 26.0& 51.9& {\bf 54.9}& {\bf 48.2}\\
\hline
\end{tabular}}
\end{center}
\caption{Results of SYNTHIA $\rightarrow$ Cityscapes. Per-category IoU, mIoU and mIoU$^*$ are presented, where mIoU denotes the results of 16 categories and mIoU$^*$ denotes the results of 13 categories. The results of other works are cited from their papers. The best result in each column is in {\bf BOLD}.}
\label{tab:syn}
\end{table*}

\subsection{Visualization}
To give more qualitative results of the proposed method, we provide several visualizations: (1) The semantic output of different models. (2) The confidence weight of different models. The confidence weight indicates the probability of the predictive category for each pixel. (3) The variance weight of different models. The variance weight is the probability that the predictive category after adding noises is different from the original predictive category.

\begin{figure*}
    \centering
    \includegraphics[width=1\textwidth]{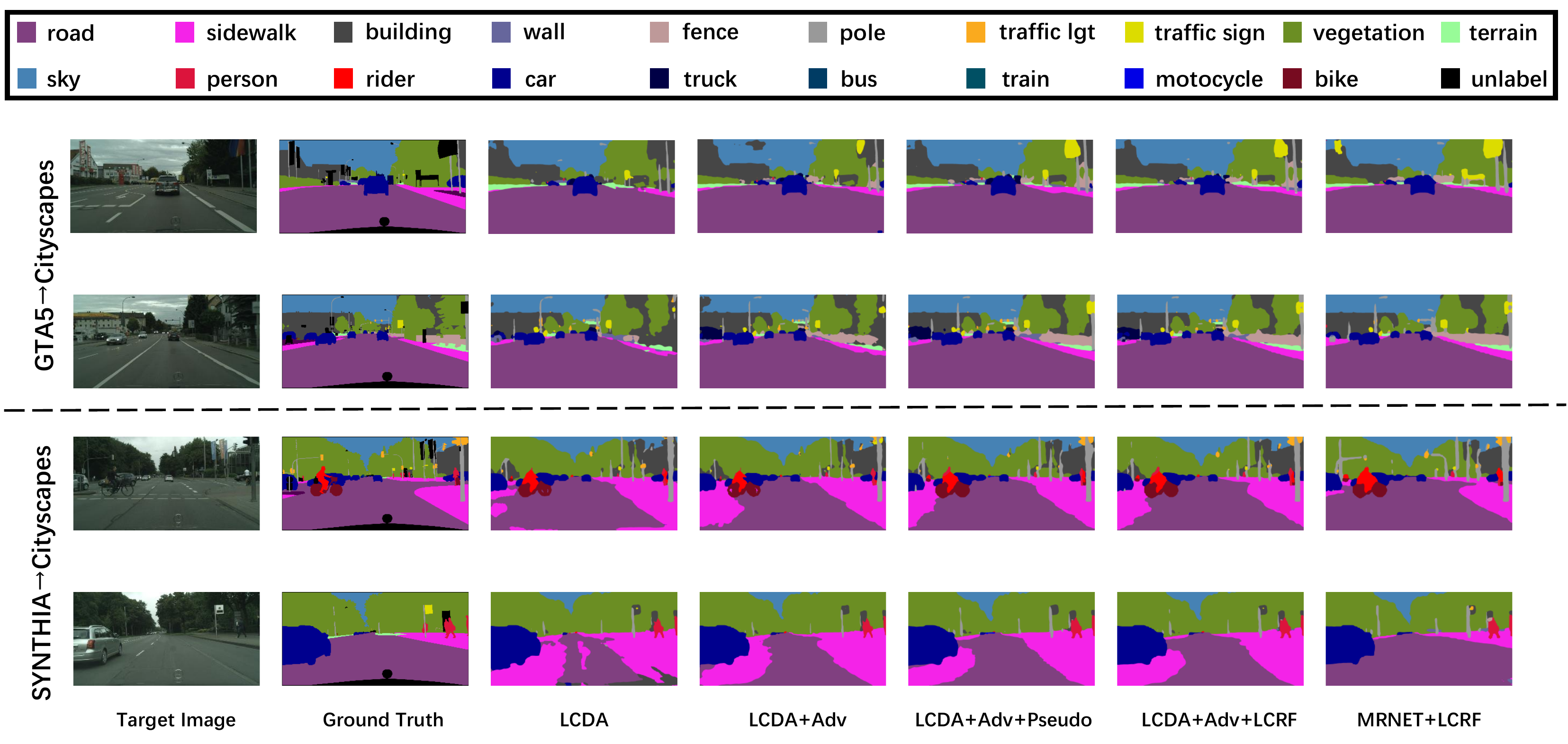}
    \caption{Visualization of semantic segmentation outputs on GTA $\rightarrow$ Cityscapes and SYNTHIA $\rightarrow$ Cityscapes. The original target image, ground truth and output of LCDA, LCDA+Adv, LCDA+Adv+Pseudo, LCDA+Adv+LCRF and MRNET+LCRF are shown.}
    \label{fig:visulization}
\end{figure*}

{\bf Semantic Output.} Visualization of semantic output is shown in Fig. \ref{fig:visulization}. It is clear that by cooperating with adversarial training, the local Lipschitzness regularization further enhances its performance. If a self-learning is introduced in the second stage, the results of an adaptive semantic segmentation model would be refined a lot. In detail, LCRF performs slightly better than Pseudo. MRNET+LCRF performs best, which avoids the error on the categories that cover a large area, such as {\tt road}, {\tt sidewalk} and {\tt sky}. It also contains more details with a small area than other methods, such as {\tt pole} and {\tt traffic sign}. It examines that a better model in stage one promotes the performance of LCRF.

\begin{figure*}
    \centering
    \includegraphics[width=1\textwidth]{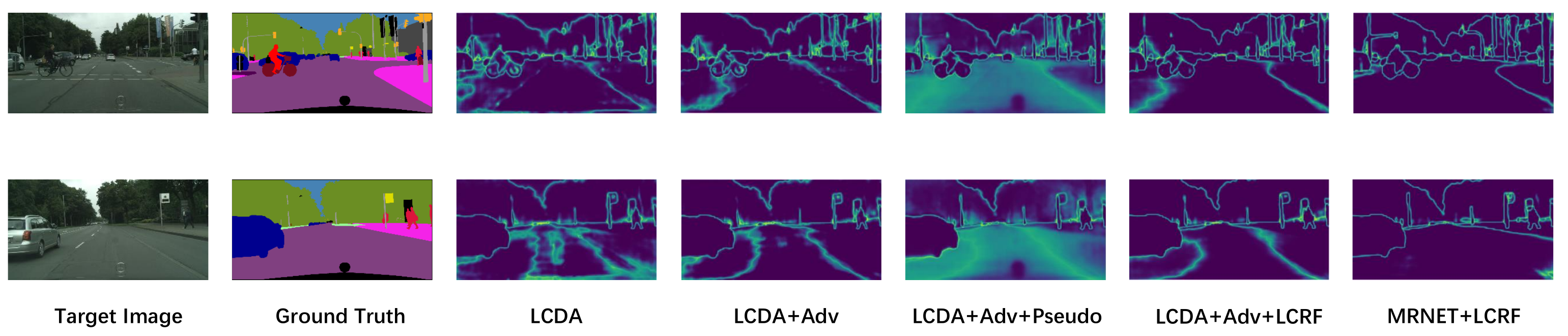}
    \caption{Visualization of confidence weights. The original target image, ground truth and confidence weights of LCDA, LCDA+Adv, LCDA+Adv+Pseudo, LCDA+Adv+LCRF and MRNET+LCRF are shown.}
    \label{fig:confidence}
\end{figure*}

{\bf Confidence Weight.} Visualization of confidence weights is shown in Fig. \ref{fig:confidence}. The confidence weight of all methods focuses on the boundary of different regions, which shows that they distinguish different categories with high confidence. An interesting discovery is that the highlighted lines of the best model, i.e., MRNET+LCRF, are thin. It means that MRNET+LCRF only gives high confidence on the edge of different categories. This phenomenon indicates that except for the edges, a good adaptive semantic segmentation method should give low confidence in other areas to avoid ambiguous predictions. However, comparing LCDA+Adv+Pseudo with LCDA+Adv+LCRF, it is clear that pseudo-label learning with manual threshold causes more ambiguous predictions. The difference verifies the robustness of the proposed dynamic self-learning.

\begin{figure}
    \centering
    \includegraphics[width=1\columnwidth]{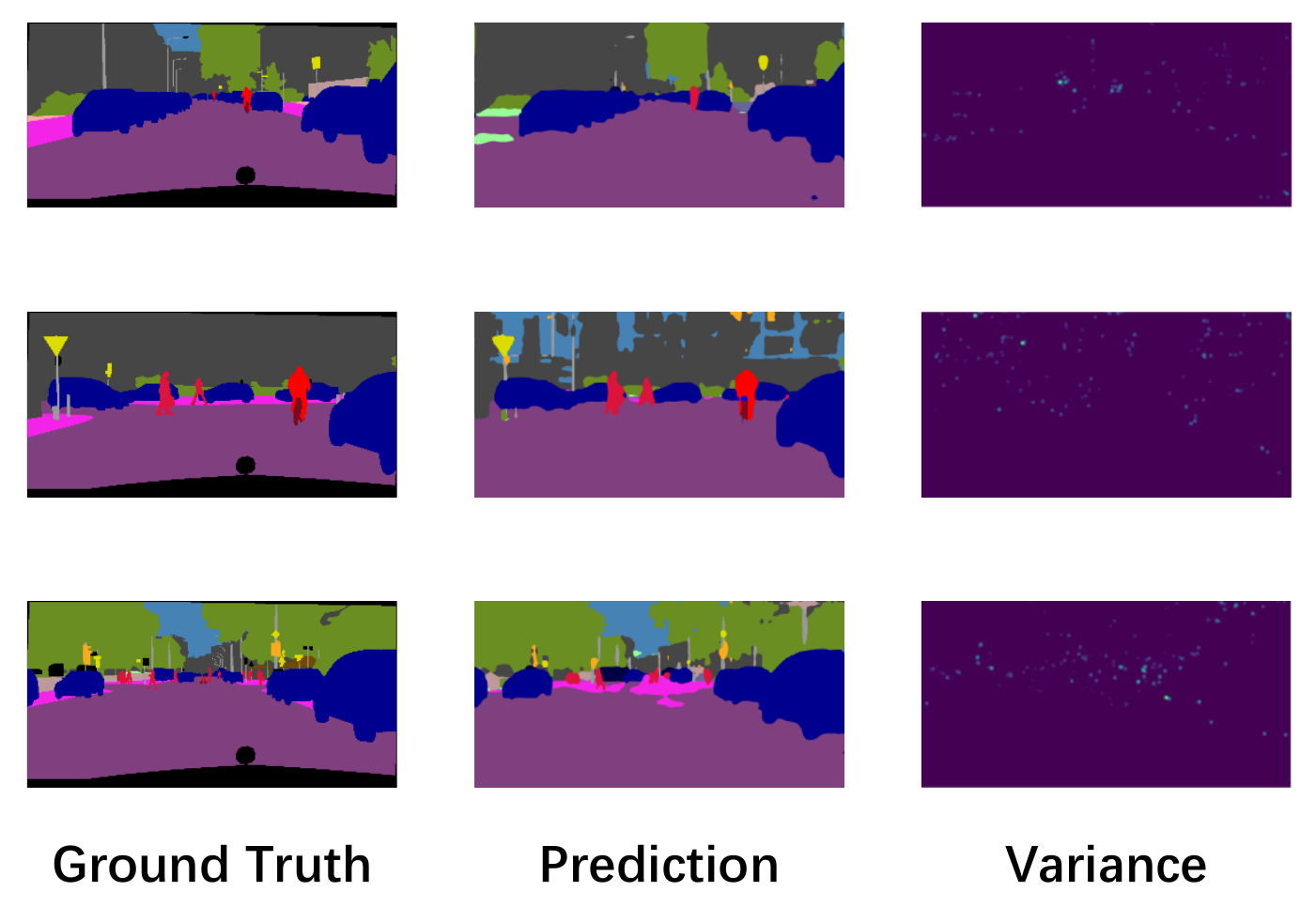}
    \caption{Visualization of variance weights. The original target image, semantic segmentation outputs and variance weights are shown.}
    \label{fig:variance}
\end{figure}

{\bf Variance Weight.} Visualization of variance weights is shown in Fig. \ref{fig:variance}. We randomly sample several variance weights of images in the Cityscapes's validation set. Large variance weights usually appear in small-scale regions. When noises are added to these regions, they are highly potential to be classified into wrong categories. These regions belong to rare categories such as {\tt traffic light} and {\tt pole}. Because annotations of these categories are scarce and large-scale regions around them tend to cover them, an adaptive semantic segmentation method is hard to generalize well on them. To solve the problem, in the first stage, the proposed local Lipschitz regularization detects such variance and tries to reduce it. In the second stage, the dynamic self-learning assigns more weights to these rare categories to ensure they would not be ignored.

\subsection{Ablation Study}
\label{ablation}

{\bf Dynamic Self-learning v.s. Manual Threshold.} To give a quantitative comparison between the proposed dynamic self-learning and pseudo-label learning with a manual threshold, we implement both of them based on LCDA+Adv and MRNET. The backbone, training setting and metric follow the description in \ref{implementation}. Results of mIoU on GTA5 $\rightarrow$ Cityscapes and mIoU* on SYNTHIA $\rightarrow$ Cityscapes are shown in Table \ref{tab:vs}.

In both GTA5 $\rightarrow$ Cityscapes and SYNTHIA $\rightarrow$ Cityscapes, LCRF performs better than pseudo-label learning with a manual threshold. LCRF gains approximate 1\% mIoU improvement in all settings. The results verify the advantage of the proposed dynamic self-learning.

\begin{table}[]
    \centering
    \begin{tabular}{|l|c|c|c|}
        \hline
         Method & GTA5$\rightarrow$Cityscapes & SYNTHIA$\rightarrow$Cityscapes \\
         \hline\hline
         LCDA+Adv+Pseudo \cite{ijcai2020-150}& 48.1 & 53.0 \\
         LCDA+Adv+LCRF & 49.1 & 54.0 \\
        \hline
        MRNET+Pseudo \cite{ijcai2020-150}& 48.3& 53.8 \\
        MRNET+LCRF & 49.9 & 54.9 \\
        \hline
    \end{tabular}
    \vspace{2mm}
    \caption{Comparison between dynamic self-learning and pseudo-label learning with manual threshold. We compare the different self-learning mechanisms based on LCDA+Adv and MRNET. All models are tested on GTA5 $\rightarrow$ Cityscapes and SYNTHIA $\rightarrow$ Cityscapes.}
    \label{tab:vs}
\end{table}

\renewcommand\arraystretch{1.0}
\begin{table*}
\begin{center}
\resizebox{\linewidth}{8mm}{\begin{tabular}{|l|ccccccccccccccccccc|c|}
\hline
Method & Road & SW & Build & Wall & Fence & Pole & TL & TS &Veg &Terrian & Sky & PR & Rider & Car & Truck & Bus & Train & Motor & Bike & mIoU\\
\hline\hline
Source& 64.8& 21.0& 75.4& 25.1& {\bf 24.8}& 29.6& 32.2& 16.6& 82.1& 31.2& 75.7& {\bf 59.0}& 13.4& 51.7& {\bf 35.7}& 20.3& {\bf 1.1}& {\bf 29.7}& 3.9& 36.5\\
LCDA & 85.2& 25.4& 81.0& 25.8& 24.7& 33.3& 37.5& 35.8& {\bf 84.2}& 27.4& {\bf 74.1}& 58.8& {\bf 25.7}& 82.0& 33.1& 31.3& 0.1& 23.9& 32.3& 43.2\\
LCDA+Adv & 85.8& 28.6& {\bf 81.5}& {\bf 27.4}& 24.3& {\bf 34.4}& {\bf 38.5}& 37.2& 84.1& 32.9& 73.9& 58.0& 18.1& {\bf 83.5}& 32.6& {\bf 33.0}& 0.2& 24.7& 28.5& 43.5\\
LCDA+Adv+LCRF & {\bf 88.4}& {\bf 36.9}& 80.1& 25.7& 23.7& 34.0& 35.1& {\bf 49.2}& 82.4& {\bf 33.9}& 70.6& 58.6& 19.3& 82.1& 27.9& 28.9& 0.0& 27.0& {\bf 43.9}& {\bf 44.6}\\
\hline
\end{tabular}}
\end{center}
\caption{Results of GTA5 $\rightarrow$ Cityscapes based on the single-head backbone. Both per-category IoU and mIoU are presented. The results of other works are cited from their papers. The best result in each column is in {\bf BOLD}.}
\label{tab:single}
\end{table*}

{\bf Architecture Agnostic.} As shown in Table \ref{tab:gta} and \ref{tab:syn}, RPLUE \cite{zheng2021rectifying} achieves comparable performance with MRNET+LCRF. RPLUE also deploys a dynamic self-training based on uncertainty estimation to avoid noisy pseudo labels. However, a limitation of it is that RPLUE requires two semantic heads to estimate uncertainty, which heavily decreases its application scope. For example, if a source-only model only includes a single semantic head, RPLUE cannot apply to it. On the contrary, our method has no requirement of network architecture such that it can apply to any semantic segmentation method. To verify the architecture agnostic, we implement LCDA, LCDA+Adv and LCDA+Adv+LCRF based on a single-head Deeplab-v2. It also adopts ResNet101 as the backbone, however, the auxiliary semantic head is not included. Both $\lambda_1$ and $\lambda_2$ are set to 1. Other settings follow the training settings in \ref{implementation}. All models are tested on GTA5 $\rightarrow$ Cityscapes.

Results of single-head models are shown in Table \ref{tab:single}. Compared with the model without adaptation, LCDA achieves 43.2\% mIoU, which gains 6.7\% mIoU improvement. By adding adversarial training, LCDA+Adv gains 0.3\% mIoU improvement over LCDA. LCDA+Adv+LCRF arrives at the best performance, i.e., 44.6\% mIoU. The results demonstrate that LCDA and LCRF perform well on single-head models. Such architecture-agnostic property makes our method practical for more application scopes.

{\bf Adaptability.} Another advantage of our method is that LCDA and LCRF can cooperate with other adaptive semantic segmentation methods to further boost the performance. To verify the adaptability of LCDA, we compare AdaptSeg \cite{tsai2018learning} with AdaptSeg+LCDA that adds (\ref{eqlocal}) to its objective function. Similarly, to show the adaptability of LCRF, we compare MRNET \cite{ijcai2020-150} with MRNET+LCRF that adopts the proposed dynamic self-learning to MRNET. The results of mIoU on GTA5 $\rightarrow$ Cityscapes and mIoU* on SYNTHIA $\rightarrow$ Cityscapes are recorded.

As the results shown in Table \ref{tab:stage1adap}, LCDA helps AdaptSeg gain 2.8\% mIoU improvement on GTA5 $\rightarrow$ Cityscapes and 2.5\% mIoU improvement on SYNTHIA $\rightarrow$ Cityscapes. LCDA boosts the performance of AdaptSeg stably on both tasks. The results shown in Table \ref{tab:stage2adap} verifies the adaptability of LCRF. Specifically, compared with MRNET, MRNET+LCRF gains 4.4\% mIoU improvement on GTA5 $\rightarrow$ Cityscapes and 4.7\% mIoU improvement on SYNTHIA $\rightarrow$ Cityscapes. The adaptability of LCDA and LCRF enables them to be a flexible basic component for adaptive semantic segmentation. Other adaptive semantic segmentation methods can adopt them to arrive at better performance. 

\begin{table}[]
    \centering
    \begin{tabular}{|l|c|c|}
        \hline
         Method & GTA5$\rightarrow$Cityscapes & SYNTHIA$\rightarrow$Cityscapes \\
         \hline\hline
         AdaptSeg \cite{tsai2018learning}& 42.4 & 46.7 \\
         AdaptSeg+LCDA & 45.2 & 49.2 \\
         \hline
    \end{tabular}
    \vspace{2mm}
    \caption{Adaptability of LCDA. Results of GTA5 $\rightarrow$ Cityscapes and SYNTHIA $\rightarrow$ Cityscapes are shown.}
    \label{tab:stage1adap}
\end{table}

\begin{table}[]
    \centering
    \begin{tabular}{|l|c|c|}
        \hline
         Method & GTA5$\rightarrow$Cityscapes & SYNTHIA$\rightarrow$Cityscapes \\
         \hline\hline
         MRNET \cite{ijcai2020-150}& 45.5 & 50.2 \\
         MRNET+LCRF & 49.9 & 54.9\\
         \hline
    \end{tabular}
    \vspace{2mm}
    \caption{Adaptability of LCRF. Results of GTA5 $\rightarrow$ Cityscapes and SYNTHIA $\rightarrow$ Cityscapes are shown.}
    \label{tab:stage2adap}
\end{table}

{\bf Effectiveness of Each Module.} In this section, we verify the effectiveness of each module in our method. Specifically, we test three modules, i.e., adversarial training \cite{tsai2018learning}, LCDA and LCRF. Results are shown in Table \ref{tab:effect}. On both tasks, adversarial training and LCDA obtain obvious improvement over the model without adaptation. LCDA performs better than adversarial training on both tasks. If LCDA cooperates with adversarial training, the performance could be further improved. Such improvement verifies that the intra-domain knowledge explored by LCDA could provide additional information for the inter-domain alignment, e.g. adversarial training. LCRF obtains 3.9\% and 4.8\% mIoU improvement over LCDA+Adv on two tasks, respectively. The results verify the effectiveness of our proposed dynamic self-learning.

\begin{table}[]
    \centering
    \begin{tabular}{|l|c|c|c|c|}
        \hline
         Task & Adversarial Training & LCDA & LCRF & mIoU \\
         \hline\hline
         GTA5$\rightarrow$Cityscapes & $-$ & $-$ &$-$ & 37.3 \\
         GTA5$\rightarrow$Cityscapes & \checkmark &$-$  & $-$& 42.4 \\
         GTA5$\rightarrow$Cityscapes & $-$ &\checkmark  & $-$& 44.0 \\
         GTA5$\rightarrow$Cityscapes & \checkmark & \checkmark &$-$ & 45.2\\
         GTA5$\rightarrow$Cityscapes & \checkmark & \checkmark & \checkmark & {\bf 49.1}\\
         \hline
         SYNTHIA$\rightarrow$Cityscapes &$-$  & $-$ & $-$& 40.4 \\
         SYNTHIA$\rightarrow$Cityscapes & \checkmark & $-$ &$-$ &46.7\\
         SYNTHIA$\rightarrow$Cityscapes & $-$ & \checkmark &$-$ & 47.4\\
         SYNTHIA$\rightarrow$Cityscapes &\checkmark  & \checkmark & $-$&49.2 \\
         SYNTHIA$\rightarrow$Cityscapes & \checkmark &\checkmark &\checkmark&{\bf 54.0} \\
         \hline
    \end{tabular}
    \vspace{2mm}
    \caption{Ablation study of each module in the proposed method. Results of GTA5 $\rightarrow$ Cityscapes and SYNTHIA $\rightarrow$ Cityscapes are shown.}
    \label{tab:effect}
\end{table}

\section{Conclusion}
In this work, we introduce the local Lipschitz constraint to solve unsupervised adaptive semantic segmentation. The local Lipschitz constraint is utilized as the unified principle to guide a two-stage training process. The first stage involves a local Lipschitzness regularization to explore intra-domain knowledge, which avoids the unstable training caused by adversarial training. The second stage introduces a dynamic self-learning based on the local Lipschitzness to alleviate the error propagation caused by pseudo labels. The unified guidance, i.e., the local Lipschitz constraint, ensures the knowledge learned in the training process to be retentive in both stages. As a result, the proposed method achieves the competitive performance on two standard benchmarks. Further, the proposed method has excellent extensibility to cooperate with any CNN-based network architectures and other adaptive semantic segmentation methods, which shows its potential to be used as flexible basic UDA components to promotes future works.


%





\ifCLASSOPTIONcaptionsoff
  \newpage
\fi

\bibliographystyle{IEEEtran}
\bibliography{lcda}

\end{document}